%% file: main.tex
\documentclass[10pt,twocolumn,letterpaper]{article}

\usepackage[pagenumbers]{cvpr} 

\input{preamble}

\definecolor{cvprblue}{rgb}{0.21,0.49,0.74}
\definecolor{linkred}{rgb}{0.8,0.16,0.16}  
\usepackage[pagebackref,breaklinks,colorlinks,citecolor=cvprblue,linkcolor=linkred]{hyperref}

\def\paperID{4214} 
\def\confName{CVPR}
\def\confYear{2024}

\title{\textit{iFusion}: Inverting Diffusion for Pose-Free Reconstruction from Sparse Views}

\author{Chin-Hsuan Wu$^{*1}$\quad Yen-Chun Chen$^{2}$\quad Bolivar Solarte$^{1}$\quad Lu Yuan$^{2}$\quad Min Sun$^{1,3}$\\
$^{1}$National Tsing Hua University \qquad
$^{2}$Microsoft  \qquad
$^{3}$Amazon \\
{\small \tt \{chinhsuanwu, enrique.solarte.nthu\}@gapp.nthu.edu.tw} \\
{\small \tt \{yen-chun.chen, luyuan\}@microsoft.com}
\quad
{\small \tt sunmin@ee.nthu.edu.tw}
\\
{ \href{https://chinhsuanwu.github.io/ifusion}{chinhsuanwu.github.io/ifusion}}
}

\begin{document}
\input{fig/banner}
\myfootnote{*}{Part of this work was done as a research intern at Microsoft.}
\input{sec/0_abstract}
\input{sec/1_intro}
\input{sec/2_prelim}
\input{sec/3_method}

\input{sec/4_exp}
\input{sec/5_related}
\input{table/exp_ab_nvs}
\input{table/exp_ab_recon}
\input{sec/6_conclusion}

\input{sec/7_ack}

{
    \small
    \bibliographystyle{ieeenat_fullname}
    \bibliography{main}
}

\clearpage
\appendix

\section*{Appendix}
We provide implementation details including hyper-parameters and dataset, more qualitative examples, and limitations and future directions as appendices.

\input{sec/a_impl}
\input{sec/b_qual}
\input{sec/c_limitation}
\input{fig/supp_pose}
\input{fig/supp_recon}


\end{document}

%% file: preamble.tex
\usepackage{ifthen}
\usepackage{color}
\usepackage{bigstrut}
\usepackage{multirow}
\usepackage{mathtools}
\usepackage[normalem]{ulem}
\usepackage{tabularx}

\definecolor{frenchblue}{rgb}{0.0, 0.45, 0.73}
\definecolor{gray}{rgb}{0.5,0.5,0.5} 
\definecolor{green}{rgb}{0, 0.4, 0} 
\definecolor{orange}{rgb}{1, 0.5, 0} 	
\definecolor{mahogany}{rgb}{0.75, 0.25, 0.0}
\definecolor{purple}{rgb}{0.6, 0, 0.6}
\definecolor{darkgreen}{rgb}{0, 0.4, 0.4} 
\definecolor{blue}{rgb}{0.0, 0.45, 0.73}
\definecolor{teal}{rgb}{0.0, 0.5, 0.5}
\definecolor{red}{rgb}{1.0, 0, 0}

\newboolean{revising}
\setboolean{revising}{false}
\ifthenelse{\boolean{revising}}
{

    \newcommand{\todo}[1]{{\textcolor{red}{TODO: #1}}}
    \newcommand{\key}[1]{{\textcolor{orange}{KEY: #1}}}
    \newcommand{\yenchun}[1]{{\color{red}{[YC: #1]}}}
    \newcommand{\justincomment}[1]{\textcolor{blue}{[justin:] #1}}
    \newcommand{\kikecomment}[1]{\textcolor{teal}{[kike:] #1}}
 }{

    \newcommand{\todo}[1]{}
    \newcommand{\key}[1]{}
    \newcommand{\yenchun}[1]{}
    \newcommand{\justincomment}[1]{}
    \newcommand{\kikecomment}[1]{}
}

\newcommand{\ours}{\textit{iFusion}\xspace}
\newcommand{\myfootnote}[2]{{%
  \let\thempfn\relax
  \footnotetext[0]{$^#1$#2}
}}

\makeatletter
\def\NAT@spacechar{~}
\makeatother

%% file: fig/banner.tex
\twocolumn[{
\maketitle
\begin{center}
    \vspace{-6mm}
    \captionsetup{type=figure}
    \includegraphics[clip, trim=0cm 20cm 13cm 0cm,width=0.95\linewidth]{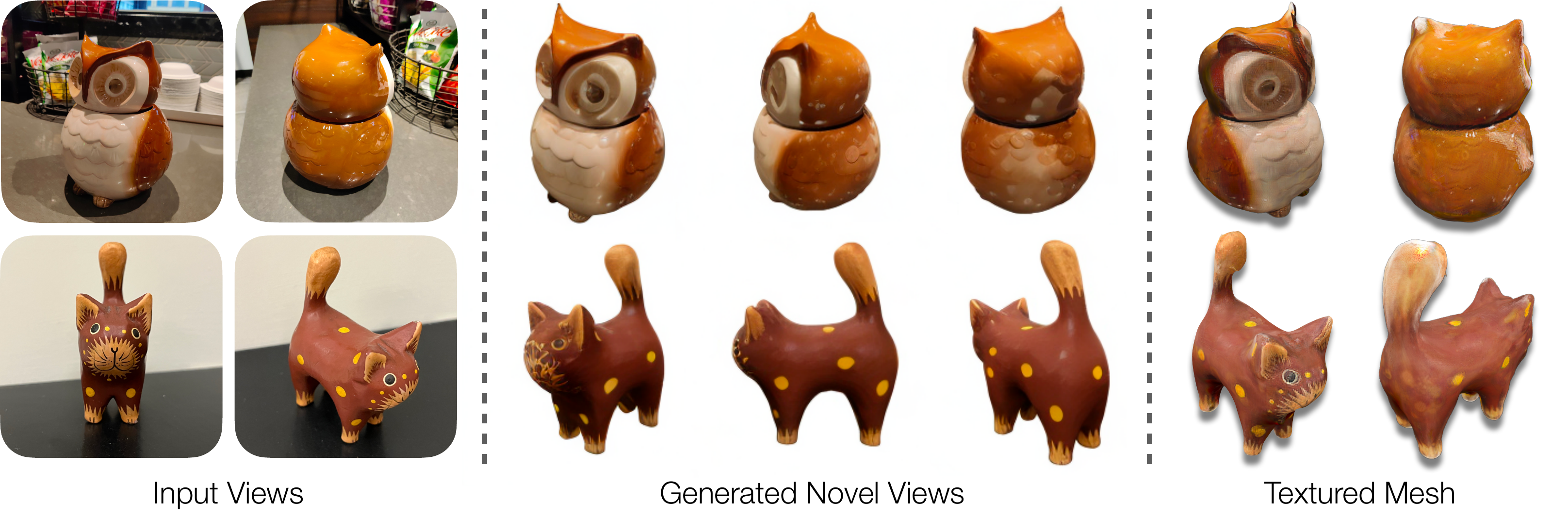}
    \vspace{-2mm}
    \caption{
        \textbf{Demonstration on real-world 3D reconstruction.} With only two casually taken photos \textbf{without camera poses}, \ours can reconstruct plausible 3D assets. The top row example is taken from DreamBooth3D~\citep{dreambooth3d}, and we took photos for the cat statue by ourselves.
    }
    \label{fig_banner}
\end{center}
}]

%% file: sec/0_abstract.tex
\begin{abstract}
\vspace{-8pt}

We present \ours, a novel 3D object reconstruction framework that requires only two views with unknown camera poses.
While single-view reconstruction yields visually appealing results, it can deviate significantly from the actual object, especially on unseen sides.
Additional views improve reconstruction fidelity but necessitate known camera poses. However, assuming the availability of pose may be unrealistic, and existing pose estimators fail in sparse-view scenarios.
To address this, we harness a pre-trained novel view synthesis diffusion model, which embeds implicit knowledge about the geometry and appearance of diverse objects.
Our strategy unfolds in three steps: (1)~We invert the diffusion model for camera pose estimation instead of synthesizing novel views.
(2)~The diffusion model is fine-tuned using provided views and estimated poses, turned into a novel view synthesizer tailored for the target object.
(3)~Leveraging registered views and the fine-tuned diffusion model, we reconstruct the 3D object.
Experiments demonstrate strong performance in both pose estimation and novel view synthesis. Moreover, \ours seamlessly integrates with various reconstruction methods and enhances them.

\end{abstract}

%% file: sec/1_intro.tex
\section{Introduction}
\label{sec:intro}

Reconstructing objects from sparse views poses a significant challenge yet holds paramount importance for various applications, including 3D content creation,~augmented reality,~virtual reality, and~robotics. Recent breakthroughs, guided by pre-trained models, have facilitated visually plausible reconstructions from a single view, without requiring the camera pose~\cite{pixelnerf, realfusion, zero123, one2345, magic123, make_it_3d, dreamgaussian}.
However, the reconstructed assets might not precisely capture the actual objects due to the inherent single-view ambiguity, \eg, the object's side opposite to the camera can only be imagined. Furthermore, multiple potential 3D structures could correspond to the same input image.

On the other hand, sparse-view methods assume the availability of an accurate camera pose for each view~\cite{dietnerf, infonerf, freenerf, sparseneus, sparsefusion, genvs, viewset_diffusion}. To meet this requirement, a Structure-from-Motion~(SfM) pre-processing, \eg, COLMAP~\cite{colmap}, is typically employed. Paradoxically, these methods demand a substantial number of images, usually more than~50 in practice, for reliable pose estimation.
Recent learning-based pose estimation~\cite{relpose, relposepp, sparsepose} and pose-free reconstruction~\cite{forge, leap} have sought to alleviate this issue. However, they still require a minimum of five input views and are primarily demonstrated on objects with simple 3D geometry or within a constrained set of object categories.
A generic framework for pose-free, sparse-view 3D reconstruction is still lacking, posing a significant obstacle to real-world applications with casually captured photos.
We hereby raise the research question: How can one utilize only \emph{extremely sparse} views \emph{without poses} while maintaining the \emph{reconstruction fidelity} of diverse objects?

The key is a sparse-view pose estimator. Our motivation stems from a recent novel view synthesis diffusion model, namely Zero123~\citep{zero123}, which is pre-trained on the most extensive 3D object dataset to date~\citep{objaverse}. Given a reference view image, Zero123 can generate a novel view~(query view) from a specified pose~(\cref{fig_teaser},~left). This indicates that the model has learned rich prior knowledge about the geometry and appearance of diverse objects.
We thus hypothesize that it can be leveraged for pose estimation, with an intuition that a well-estimated pose fed into Zero123 will produce an image similar to the query view, and vice versa. Next, gradients may be back-propagated to optimize the pose with a proper loss function.
Following this idea, we repurpose Zero123 by inverting it to take the two views and estimate the relative camera transformation~(\cref{fig_teaser},~right). More specifically, we adopt an analysis-by-synthesis paradigm~\citep{inerf, neural_object_fitting, latentfusion} that optimizes the transformation by minimizing the difference between the denoised latent visual features, \ie, Zero123's output image feature map, and the query view's feature.
Empirically, the proposed approach achieves strong pose estimation with as few as~2 views, even outperforming existing approaches' results with~5 views.

Well-estimated poses also open up a new opportunity. Using the given views registered with poses, a mini-dataset can be constructed to further fine-tune Zero123 and customize the diffusion model for synthesizing the target object's novel views. Specifically, we can form a set of~(reference view, camera pose, query view) triplets from the given sparse views and fine-tune Zero123.
To accelerate training and prevent overfitting, we use Low-Rank Adaptaion~(LoRA)~\citep{lora} to fine-tune the diffusion model, a recognized technique for customizing diffusion models.\footnote{\href{https://github.com/cloneofsimo/lora}{https://github.com/cloneofsimo/lora}}
Experiments demonstrate that this step significantly improves novel view synthesis, achieving an average increase of~\textbf{+3.6} in PSNR across two datasets, and is beneficial to the final reconstruction.
Note that our approach shares a similar spirit with test-time training~\citep{sun2020test}, test-time adaptation~\citep{wang2020tent}, and self-training~\citep{Scudder1965probability,xie2020self}.
Like test-time training and adaptation, we align the model to the test distribution based on test inputs~(given views) but without test labels~(novel views). Analogous to self-training, we synthesize additional labels~(camera poses) using the learning model itself.
To the best of our knowledge, the above combination is new for diffusion-based 3D reconstruction.

To this end, we introduce \ours, a novel framework that reconstructs diverse 3D objects with sparse, pose-free views. First, the pose estimation is achieved by \textbf{i}nverting the Zero123 dif\textbf{Fusion} model, as described earlier. With the estimated camera pose, an object-specific improvement on Zero123's novel view synthesis capability is performed, which can be further utilized as additional reconstruction guidance. Finally, for reconstructing the 3D asset, any differentiable renderer can be plugged in, including NeRFs~\citep{nerf} and the recently proposed 3D Gaussian Splatting~\citep{gaussian}. It is noteworthy that our framework does not assume any specific reconstruction pipeline, and experimental results demonstrate that \ours is readily applicable to four different single-view reconstruction methods. Improved geometric fidelity is observed with a significant \textbf{+7.2\%} increase in volume IoU, showcasing the necessity of additional views for reliable 3D reconstruction.

\input{fig/teaser}

Our contributions are summarized as follows:
\begin{enumerate}
    \item We propose a novel camera pose estimator that significantly outperforms existing methods in terms of both accuracy and required number of input views, while being effective for diverse objects.
    \item A self-training and test-time training inspired fine-tuning stage is innovated. This stage results in a much stronger novel view synthesis diffusion model, which plays a crucial role in guiding the reconstruction process.
    \item For the first time, we escalate diffusion-based single-view reconstruction to multi-view for enhanced fidelity with merely two pose-free images.
\end{enumerate}

%% file: fig/teaser.tex
\begin{figure}[t]
    \centering
    \small
    \includegraphics[clip, trim=0cm 6.5cm 21cm 0cm,width=\columnwidth]{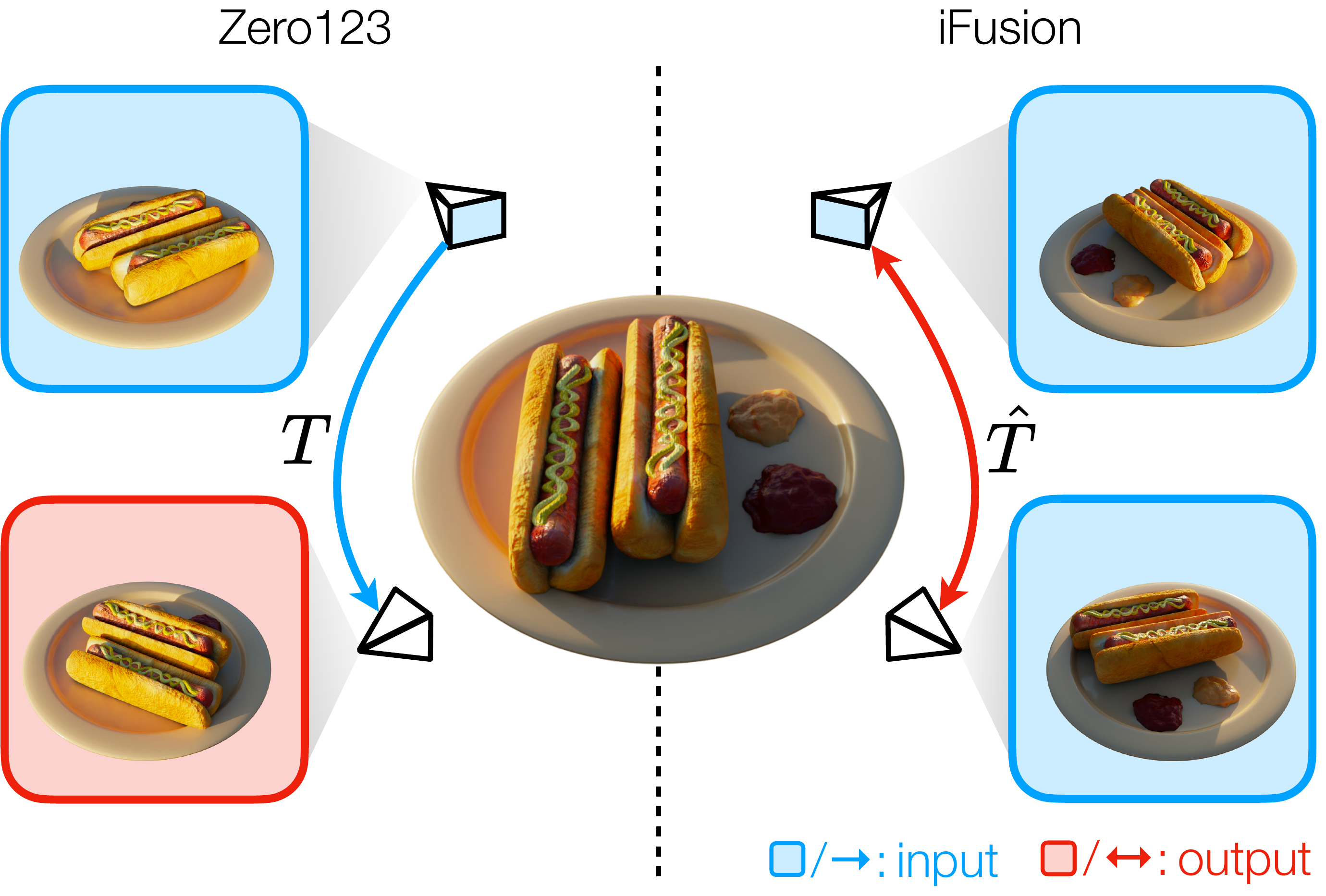}
    \vspace{-6mm}
    \footnotesize
    \caption{
        \textbf{Zero123 \vs \ours.} Unlike Zero123~\cite{zero123}~(left), which synthesizes an object's novel view given an image and a transformation $T$, \ours~(right) instead optimizes an unknown relative transformation $\hat{T}$ from two given views.
    }
    \vspace{-4pt}
    \label{fig_teaser}
\end{figure}

%% file: sec/2_prelim.tex
\section{Preliminary}
\label{sec:prelim}

\ours repurposes a novel view synthesizing diffusion model for camera pose prediction. To prepare readers with the necessary backgrounds, we briefly introduce the basics of Diffusion Models~(DM) and how they can be used for novel view synthesis. Next, we summarize a recently popular approach to utilize DM for 3D reconstruction, which we integrate into \ours to allow reconstruction.

\vspace{-4pt}
\paragraph{Diffusion Models}\label{sec:prelim:zero123}
Diffusion models~\cite{ddpm, dpm, ncsn} are a class of deep generative models that has become the mainstream approach for high-fidelity visual synthesis.
In image generation, they work by ``diffusing" an image by adding noise over repeated steps, and then a deep neural network is trained to predict the applied step-wise noise from a corrupted image.
This allows the reversion of the diffusion process, thus an image can be generated from a random noise by iterative denoising using the trained noise predicting network.
More specifically, \citet{ddpm} formulated the diffusion process in the following analytical form:
\begin{equation}\label{eq_prelim_x_t}
x_t = \sqrt{\alpha_t}x_0 + \sqrt{1 - \alpha_t}\epsilon, \quad t \in [0, 1, \dots, \mathcal{T}],
\end{equation}
where $\epsilon \sim \mathcal{N}(0, 1)$ denotes the Gaussian noise and hyper-parameter~$\alpha_t$ denotes the noise schedule.
For the reverse process, the noise predictor is denoted as~$\epsilon_\theta(x_t, t)$, where $\theta$ is the set of trainable parameters.
Instead of directly modeling the RGB pixel values~$x$, a widely used diffusion model, Stable Diffusion~(SD),\footnote{\href{https://github.com/CompVis/stable-diffusion}{https://github.com/CompVis/stable-diffusion}} applies the Latent Diffusion Model~(LDM)~\cite{ldm} to model the latent feature maps~$z$. The encoding and reconstruction of images is done via a pre-trained VQ-VAE:~$z = \mathcal{E}(x)$, and~$x = \mathcal{D}(z)$.
Moreover, DM may optionally take conditional inputs~$c$, \eg, texts,~bounding box layouts, and~depth maps. For instance, the standalone SD takes texts as the condition~$c$ and enables text-to-image generation~(T2I). Formally, the training loss of the prediction network can be written as:
\begin{equation} \label{eq_ldm}
\mathcal{L}(x, c) = \mathbb{E}_{z,\epsilon,t}\left [ \left \| \epsilon - \epsilon_\theta(z_t, t, c) \right \|_2^2 \right ],
\end{equation}
where $ \| \cdot \|_2$ denotes the L2 norm.

\vspace{-4pt}
\paragraph{Diffusion Models for Novel View Synthesis}
The original Stable Diffusion was trained on web-scale image-text pairs\footnote{\href{https://laion.ai/blog/laion-aesthetics/}{https://laion.ai/blog/laion-aesthetics/}} for text-to-image generation. Recently, \citet{zero123} proposed Zero123 to further fine-tune SD on Objaverse~\citep{objaverse}, a large-scale 3D assets dataset, for object-centric novel view synthesis. Given an image at the reference viewpoint~$x^r$ and the \textbf{r}eference-to-\textbf{q}uery \textbf{t}ransformation~$T_{r \rightarrow q} \in $ SE(3), the model synthesizes the desired query view~$x^q$ with condition~$c(x^r, T_{r \rightarrow q})$. This is formulated as a DM and shares the same training objective as~\cref{eq_ldm}.

\vspace{-4pt}
\paragraph{3D Reconstruction via Score Distillation Sampling}\label{sec:prelim:sds}
Recent studies~\cite{clip_mesh, dreamfields, dreamfusion, sjc} indicated that large-scale pre-trained 2D vision models~\cite{clip, imagen, ldm} implicitly encapsulate rich 3D geometric prior. Notably, DreamFusion~\cite{dreamfusion} introduced the Score Distillation Sampling~(SDS) to facilitate 3D generation guided by a pre-trained 2D DM. Let $x = \mathcal{R}_\psi(T)$ be the rendered image at viewpoint~$T \in$ SE(3), where $\mathcal{R}$ is a differentiable renderer parameterized by~$\psi$, \eg, Neural Radiance Fields~(NeRFs)~\cite{nerf} or 3D Gaussian Splatting~\cite{gaussian}. Given a denoising network~$\epsilon_\theta$, SDS optimizes the renderer~$\psi$ by minimizing the residuals between the predicted noise and the added noise, thereby producing the gradients:
\begin{equation}\label{eq_prelim_sds}
\nabla_\psi \mathcal{L}_{SDS}(x, c) = \mathbb{E}_{z, \epsilon, t} \left [ (\epsilon_\theta(z_t,t,c) - \epsilon)\frac{\partial z}{\partial \psi} \right ].
\end{equation}

%% file: sec/3_method.tex
\input{fig/pipeline}
\section{Method}
\label{sec:method}

\Cref{fig_pipeline} presents an overview of the \ours framework. The key of our pose-free reconstruction framework is the sparse-view pose estimator shown in \cref{fig_pipeline}~(a). By inverting the diffusion model, accurate poses can be estimated.
Next, the registered views are leveraged to customized the novel view synthesis model for the target object as in \cref{fig_pipeline}~(b). Finally, 3D reconstruction can be done using the registered views, and the customized diffusion model serves as the guidance, shown in \cref{fig_pipeline}~(c).

\subsection{Diffusion as a Pose Estimator} \label{sec:method:pose}
The goal is to recover the relative camera pose~$T_{r \rightarrow q}$ from a reference view~$x^r$ to the query view~$x^q$, leveraging the pre-trained diffusion model~$\epsilon_\theta$. Intuitively, a model trained for a task involving camera poses could potentially be used in reverse: to retrieve or estimate the camera pose from given inputs, as evident in~\citet{inerf, neural_object_fitting, latentfusion}. Hence, rather than optimizing DM parameters~$\theta$ to reconstruct $x^q$ given $c(x^r, T_{r \rightarrow q})$ as in the training stage shown in~\cref{eq_ldm}, we solve the inverse problem by freezing $\theta$ and optimizing $\hat{T}_{r \rightarrow q}$ to reconstruct $x^q$:
\begin{equation} \label{eq_pose1}
    \hat{T}_{r \rightarrow q} = \mathop{\text{argmin}}\limits_{T \in \text{SE(3)}} \mathcal{L}(x^q, c(x^r, T)).
\end{equation}

To minimize~\cref{eq_pose1}, we query a view in its latent space~$z_t \sim \mathcal{E}(x^q)$ using~\cref{eq_prelim_x_t}, followed by denoising $z_t$ to~$\hat{z}_{t-1}$ conditioned on~$c(x^r, \hat{T}_{r \rightarrow q})$. Finally, we compute the residuals for backpropagation of the transformation's gradient~$\nabla\hat{T}_{r \rightarrow q}$. To ensure that the estimated pose~$\hat{T}_{r \rightarrow q}$ continue to lie on the SE(3) manifold during the gradient-based optimization, we parameterize the pose~$T_{r \rightarrow q}=\exp(\xi)$, where $\xi \in \mathbb{R}^6$ is the twist coordinates of the Lie algebra~$\mathfrak{se}(3)$ associated with the Lie group~SE(3)~\cite{lietheory}. Therefore, we reformulate~\cref{eq_pose1} as follows:
\begin{equation} \label{eq_pose}
    \hat{\xi}_{r \rightarrow q} = \mathop{\text{argmin}}\limits_{\xi \in \mathfrak{se}(3)} \mathcal{L}(x^q, c(x^r, \exp(\xi))).
\end{equation}
Note that~\cref{eq_pose} can further be constrained by the inverse transformation defined by the same vector representation, \ie, $T_{q \rightarrow r}=\exp(-\xi)$. We therefore obtain:
\begin{equation} \label{eq_pose2}
\begin{aligned}
    \hat{\xi}_{r \rightarrow q} = \mathop{\text{argmin}}\limits_{\xi \in \mathfrak{se}(3)} \; & \mathcal{L}(x^q, c(x^r, \exp(\xi))) \\
    + \; &\mathcal{L}(x^r, c(x^q, \exp(-\xi))).
\end{aligned}
\end{equation}

In practice, we initialize our optimization from four distinct canonical poses relative to the reference view, \ie, front,~left,~right, and~back, designated as $T_0$. This helps reduce the possibility of stucking at a local minima during the optimization. The final estimated camera pose can be denoted as follows:
\begin{equation} \label{eq_pose3} \hat{T}_{r \rightarrow q} = T_0 \cdot \exp(\hat{\xi}_{r \rightarrow q}).
\end{equation}

Furthermore, taking inspiration from~\citet{dreamtime}, instead of sampling the timestep~$t$ from a uniform distribution as in training, we linearly decrease $t$. This adjustment aligns with diffusion models' coarse-to-fine progressive optimization and has been empirically observed to lead to more stable optimization.

\input{fig/exp_pose}

\subsection{From Single-View to Multi-View} \label{sec:method:nvs}

Even with a fairly accurate estimated pose~$\hat{T}_{r \rightarrow q}$, there is still no guarantee that the diffusion model generates the pixel-exact query image~$x^q$. We propose to close the gap by further fine-tuning the DM with the given views and estimated poses.
However, due to limited training samples, naively optimizing all trainable parameters~$\theta$ is inefficient and may jeopardize the pre-trained model. To this end, we incorporate LoRA~\cite{lora}, injecting thin trainable layers~$\phi$ to the attention module in the U-Net~$\epsilon_\theta$ while freezing the pre-trained $\theta$. The objective in~\cref{eq_ldm} is reformulated as follows:
\begin{equation} \label{eq_ldm_lora}
\mathcal{L}_\phi(x, c) = \mathbb{E}_{z,\epsilon,t}\left [ \left \| \epsilon - \epsilon_{\theta, \phi}(z_t, t, c) \right \|_2^2 \right ],
\end{equation}
where $(x, c) \in \left \{\left (x^q, (x^r, \hat{T}_{r \rightarrow q}) \right ), \left (x^r, (x^q, \hat{T}_{q \rightarrow r}) \right ) \right \}$.
In other words, the fine-tuning process adapts the DM to generate the query view~$x^q$ from condition~$c(x^r, \hat{T}_{r \rightarrow q})$, and vice versa, for a specific object.
Empirically, this LoRA fine-tuning effectively customize the DM to generate novel views different from~$x^r$ and~$x^q$ of the target object, despite the small number of training samples and~parameters~$\phi$, and the inherent noise from the estimated poses.


While the original Zero123 only conditions on a single view, we have multiple images available along with their relative transformations in a sparse-view setting.\footnote{We mainly formulate the two-view setting~($x^r$ and~$x^q$). Multi-view settings are achieved via treating all distinct image pairs as query-reference pairs and estimating the pose transform for each pair.} This raises the question: How can we better utilize these additional views for improved generation quality? To address this, we employ a simple stochastic conditioning strategy inspired by~\citet{3dim}.
The key concept is that all given views should collectively shape the final output. More specifically, we randomly sample a registered view as the input condition at each denoising timestep. Empirically, this stochastic multi-view conditioning~(MVC) significantly improves the novel view synthesis results compared to naively using the nearest view as the condition. Moreover, the final reconstruction quality is also improved.

\subsection{From Sparse Views to 3D Reconstruction} \label{sec:method:recon}
There are two primary lines of existing literature for 3D object reconstruction via diffusion, namely image-based reconstruction~\cite{one2345, syncdreamer} and SDS-based generation~\cite{dreamfusion, magic3d, dreamgaussian, magic123}. To integrate our proposed technique with the image-based approaches, we may simply generate multi-view images using the fine-tuned model obtained from~\cref{eq_ldm_lora} with stochastic multi-view conditioning outlined in~\cref{sec:method:nvs}, and then feed them as the training data to the differentiable renderer, \eg, NeRF~\cite{nerf} and NeuS~\cite{neus}.
For SDS-based methods, in addition to~\cref{eq_prelim_sds}, we further incorporate the reconstruction loss on the registered input views:
\begin{equation}\label{eq_recon_sds_obj}
\mathcal{L}_{rec} = \left \| x - \mathcal{R}_\psi(\hat{T}) \right \|^2_2,
\end{equation}
where $x$ is the input image and $\mathcal{R}_\psi(\hat{T})$ is the rendered view from viewpoint $\hat{T}$ acquired from~\cref{eq_pose3}. The final objective is the weighted sum of $\mathcal{L}_{rec}$ and~$\mathcal{L}_{SDS}$. For above steps, the LoRA model and MVC are also employed.

%% file: fig/pipeline.tex
\begin{figure*}[t]
    \small
    \centering
    \footnotesize
    \includegraphics[clip, trim=0cm 21cm 13.2cm 0cm, width=\linewidth]{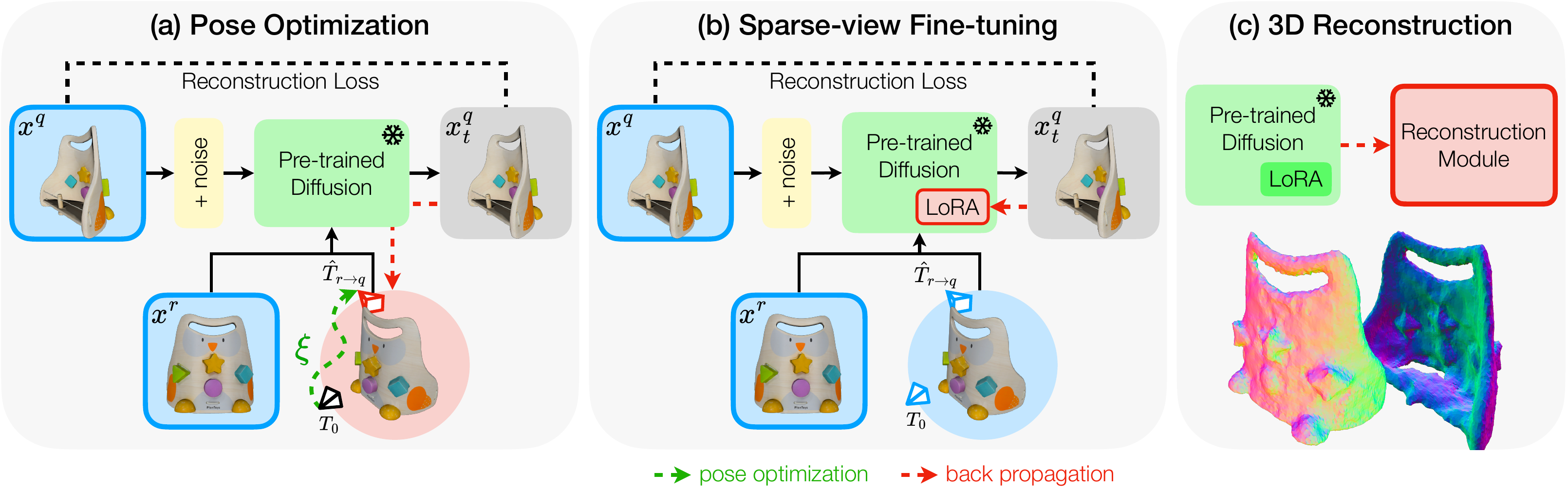}
    \vspace{-5mm}
    \caption{
        \textbf{iFusion framework.} (a)~Given as few as two pose-free images~$(x^r, x^q)$, we estimate the pose~$\hat{T}_{r \rightarrow q}$ from~${T}_0$ to optimally reconstruct the input view through the frozen diffusion model.
        (b)~Based on $\hat{T}_{r \rightarrow q}$, we efficiently fine-tune the diffusion model by LoRA~\cite{lora} to customize the model to synthesize novel views of the given object with enhanced fidelity. (c)~Conditioned on $\hat{T}_{r \rightarrow q}$ and the refined diffusion model, we optimize a reconstruction module to perform sparse view 3D reconstruction.
    }
    \label{fig_pipeline}
\end{figure*}

%% file: fig/exp_pose.tex
\begin{figure*}[t]
    \small
    \centering
    \footnotesize
    \includegraphics[clip, trim=0cm 23cm 0cm 0cm, width=0.9\linewidth]{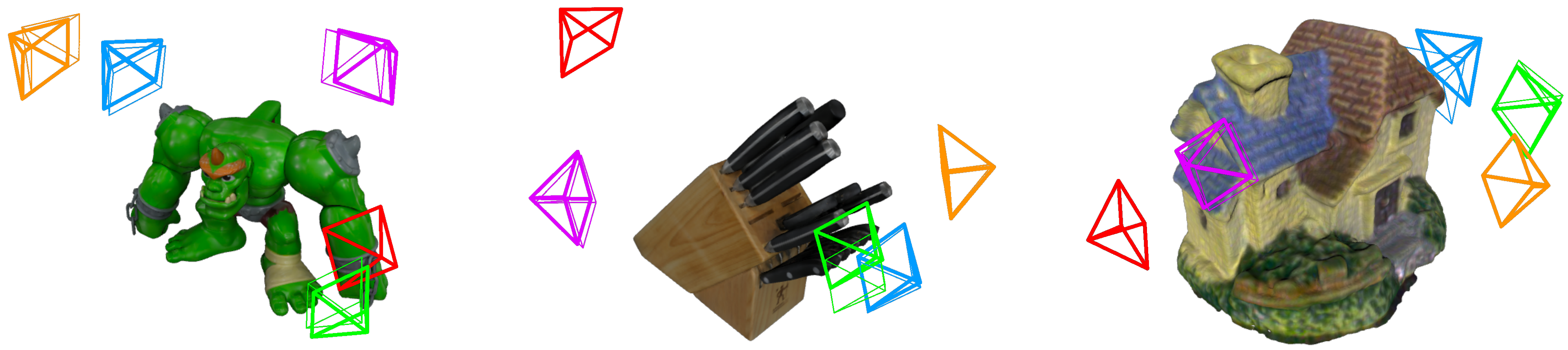}
    \vspace{-8pt}
    \caption{\textbf{Qualitative results on pose estimation.} We visualize the predicted poses~(thin) alongside the ground truth~(bold), using the same color, while the \textcolor{red}{reference views} are plotted in red. \ours accurately predicts poses even on the opposite side of the \textcolor{red}{reference view}~(red), emphasizing its effectiveness in leveraging the strong prior knowledge embedded in Zero123~\cite{zero123}.
    }
    \label{fig_exp_pose}
\end{figure*}

%% file: sec/4_exp.tex
\section{Experiments}
\label{sec:exp}

\input{fig/exp_nvs}

\subsection{Experimental Setup}

\paragraph{Datasets} We conduct experiments using two publicly available object datasets: Google Scanned Object~(GSO)~\cite{gso} and~OmniObject3D~(OO3D)~\cite{omniobject3d}. We sample $70$ instances from each dataset, randomly synthesizing camera poses and rendering observation views. For pose estimation experiments, we render five views per object, accumulating $1,400$ views in total with their corresponding camera poses for each dataset.
Regarding novel view synthesis and 3D reconstruction experiments, we sample two views from the rendered five with the largest parallax motion around the object to minimize the overlapping between views.

\paragraph{Experiments and Metrics} 
We evaluate our proposed framework on pose estimation,~novel view synthesis, and~3D reconstruction.
For pose estimation, we report the median error in rotation and translation along with a recall evaluation with a $5^\circ$ threshold for both, \ie, we consider a true positive only when both rotation and translation errors are within the threshold. Recall results are reported in percentage.
Following~\citet{nerf, zero123}, we adopt the standard metrics PSNR,~SSIM, and~LPIPS to evaluate novel view synthesis results.
For 3D reconstruction, we report Chamfer Distances and volumetric IoU between ground truth shapes and reconstructed ones.

\subsection{Experimental Result}

\paragraph{Pose Estimation} 
We compare our proposed \ours with RelPose++~\cite{relposepp} and FORGE~\cite{forge} for pose estimation given two views of each object. Quantitative and qualitative results are depicted in~\cref{exp_pose} and~\cref{fig_exp_pose}, respectively.
\Cref{exp_pose} verifies the effectiveness of our proposed solution over the baselines with significant improvements for all metrics. We found that by leveraging the diffusion model~\cite{zero123}, \ours excels at handling diverse objects thanks to its strong prior knowledge learned during pre-training, whereas RelPose++ and FORGE fall short due to their smaller training dataset with limited object diversity.
Based on the qualitative results presented in~\cref{fig_exp_pose}, we corroborate the benefits of our proposed solution in estimating the pose between two given views. We consistently find that our solution estimates accurate camera poses even with minimal overlapping. This is evident in~\cref{fig_exp_pose}, where all samples show several cameras on the opposite sides to the camera reference~(red camera) and \ours still achieves accurate estimations.
Notably, COLMAP~\cite{colmap} cannot serve as a baseline in our evaluation due to the structural limitations of Structure-from-Motion, which requires a large number of views for optimization.

\input{table/exp_pose}

\input{table/exp_nvs}
\paragraph{Novel View Synthesis}
\Cref{exp_nvs} shows our novel view synthesis comparison against 2D-based Zero123\footnote{By default, we use Zero123-XL for all modules that require Zero123.} and 3D-based methods, \ie, FORGE and LEAP~\cite{leap}. It is observed that both the 3D-based methods do not perform well under extremely few-view scenarios.
Moreover, \ours significantly outperforms all methods on all metrics.
\Cref{fig_exp_nvs} includes qualitative examples to demonstrate \ours's advantage in novel view synthesis. We observe that images generated by Zero123, although mostly visually plausible, do not faithfully represent the actual objects, especially those with complex geometry.
In contrast, our \ours improves novel views' image fidelity by conditioning on an additional pose-free view.

\input{fig/exp_recon}
\input{table/exp_recon}

\paragraph{3D Reconstruction} 
We showcase the efficacy of the \ours framework in 3D reconstruction by integrating it with various existing reconstruction methods.
Specifically, One-2-3-45~\cite{one2345} represents image-based methods, which directly regresses SDFs from the generated multi-view images;
on the other hand, Zero123-SDS~\cite{zero123},~Magic123~\cite{magic123}, and~DreamGaussian~\cite{dreamgaussian} are SDS-based approahces.
For completeness, Zero123-SDS trains Instant-NGP~\citep{mueller2022instant} via Zero123-guided SDS. Magic123 combines Zero123 and SD for improved quality.\footnote{The implementations of Zero123-SDS and Magic123 are adopted from threestudio: \href{https://github.com/threestudio-project/threestudio}{https://github.com/threestudio-project/threestudio}.}
DreamGaussian leverages the recent 3D Gaussian Splatting renderer~\cite{gaussian}.
As illustrated in~\cref{exp_recon} and~\cref{fig_exp_recon}, the incorporation of \ours enhances the performance of all reconstruction modules by a large margin.
In addition, \ours clearly outperforms other none-optimization-based methods Point-E~\citep{point_e} and Shape-E~\citep{shap_e}, which are trained on a large-scale private dataset. To conclude, when faithful reconstruction is desired, \ours is extremely beneficial, requiring very few additional view that can be casually captured without knowing the camera poses.


\subsection{Ablation Study}

\vspace{-2pt}
\paragraph{Pose Estimation} We first validate whether the use of more poses for initialization, namely $T_0$ in \cref{eq_pose3}, leads to more accurate camera pose estimation, and it is confirmed in~\cref{exp_ab_pose}.
The reported computation time was measured on a single Nvidia 3090 GPU.
Based on \cref{exp_ab_pose}, we employed $n = 4$ initial poses for a better trade-off between speed and accuracy for all experiments unless otherwise specified.
Additionally, we observed that linearly annealing the timestep~$t$ lead to significantly more accurate pose estimation, as demonstrated in~\cref{exp_ab_pose_t}.

\vspace{-10pt}
\paragraph{Sparse-view Fine-tuning} 
\cref{exp_ab_nvs} assesses the efficacy of the proposed fine-tuning stage for object-specific novel view synthesis.
Upon examining row~(a), \ie, Zero123, alongside row~(b), it is evident that the performance is boosted by incorporating the additional view and an accurately estimated pose. Row~(c) highlights the substantial improvement from the stochastic re-sampling of multi-view conditions at each timestep, providing more robust outcomes than row~(b). Moreover, the multi-view fine-tuning with LoRA in row~(d) significantly enhances performance by improving the understanding of the target object. Finally, row~(e) underscores the potential for achieving higher-quality synthesis by incorporating more views. All are achieved with self-estimated camera poses.

\vspace{-8pt}
\paragraph{3D Reconstruction}
We validate the proposed components contributing to reconstruction in~\cref{exp_ab_recon}, using DreamGaussian as the reconstruction module on the OO3D dataset. The results in rows~(a) and~(b) distinctly illustrate that adding an extra view with an estimated pose and supervising with reconstruction loss significantly enhance the single-view baseline. Incorporating stochastic multi-view conditioning~(MVC) further improves the performance, as evident in row~(c). Finally, fine-tuning via LoRA demonstrates an additional improvement in customizing the model for faithful reconstruction of the given object.

\input{table/exp_ab_pose}

%% file: fig/exp_nvs.tex
\begin{figure*}[t]
    \small
    \centering
    \footnotesize
    \includegraphics[clip, trim=0cm 2.5cm 0cm 0cm, width=\linewidth]{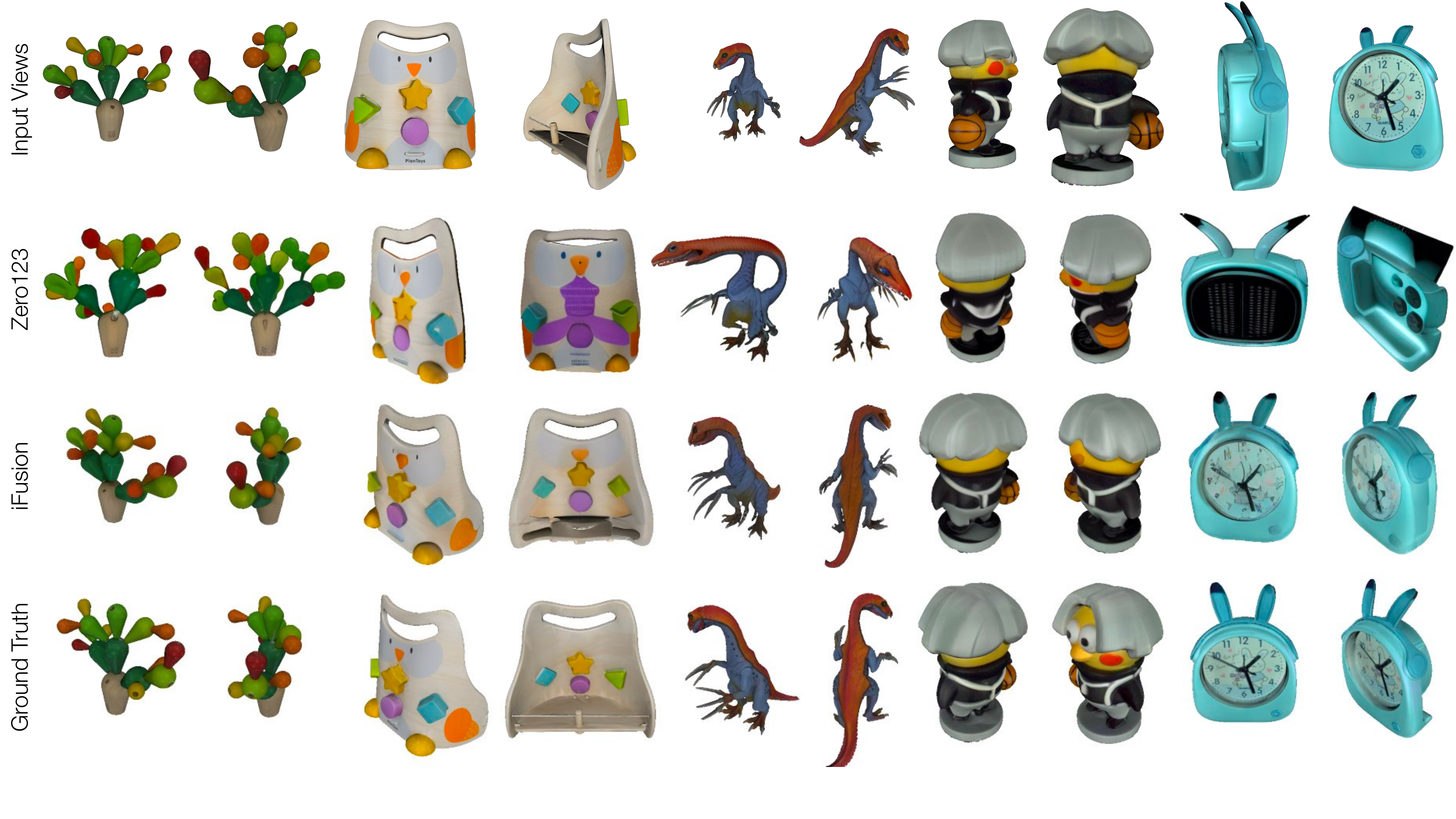}
    \caption{\textbf{Qualitative examples on novel view synthesis.} \ours takes two unposed images and Zero123~\citep{zero123} only conditions on the first view. We observe that \ours effectively leverages the additional images without camera poses and generates more faithful images.
    }
    \label{fig_exp_nvs}
\end{figure*}

%% file: table/exp_pose.tex
\begin{table}
\fontsize{8.7}{12}\selectfont
  \caption{\textbf{Evaluation results on pose estimation.} \ours achieves significant improvements for all metrics under 2 input views.}
  \label{exp_pose}
  \centering
  \begin{tabular}{llccc}
    \toprule
    
    Dataset & Method & Rot. $\downarrow$ & Trans. $\downarrow$ & Recall $\uparrow$ \\
    
    \midrule
    
    \multirow{3}{*}{GSO~\cite{gso}}
        & RelPose++~\cite{relposepp} & 109.89 & 90.58 & 0.21 \\
        & FORGE~\cite{forge} & 111.15 & 88.01 & 0.00 \\
        & \ours & \textbf{2.29} & \textbf{2.22} & \textbf{74.79} \\

    \midrule

    \multirow{3}{*}{OO3D~\cite{omniobject3d}}
        & RelPose++ & 108.83 & 90.83 & 0.00 \\
        & FORGE & 107.82 & 87.21 & 0.00 \\
        & \ours & \textbf{2.97} & \textbf{2.80} & \textbf{69.29} \\
    
    \bottomrule
  \end{tabular}
\end{table}

%% file: table/exp_nvs.tex
\begin{table}
\fontsize{9}{12}\selectfont
  \caption{\textbf{Novel view synthesis results.} \ours performed significantly better than the original Zero123 and 3D-based methods.}
  \label{exp_nvs}
  \centering
  \begin{tabular}{llccc}
    \toprule
    
    Dataset & Method & PSNR$\uparrow$ & SSIM$\uparrow$ & LPIPS$\downarrow$ \\
    
    \midrule
    
    \multirow{4}{*}{GSO~\cite{gso}}
        & FORGE~\cite{forge} & 10.45 & 0.673 & 0.449 \\
        & LEAP~\cite{leap} & 12.51 & 0.751 & 0.312 \\
        \cmidrule{2-5}
        & Zero123~\cite{zero123} & 15.40 & 0.788 & 0.184 \\
        & \ours & \textbf{18.73} & \textbf{0.836} & \textbf{0.121} \\

    \midrule

    \multirow{4}{*}{OO3D~\cite{omniobject3d}}
        & FORGE & 10.48 & 0.684 & 0.447 \\
        & LEAP & 12.63 & 0.759 & 0.305 \\
        \cmidrule{2-5}
        & Zero123 & 15.84 & 0.801 & 0.184 \\
        & \ours & \textbf{19.78} & \textbf{0.851} & \textbf{0.117} \\
    
    \bottomrule
  \end{tabular}
\end{table}

%% file: fig/exp_recon.tex
\begin{figure*}[t]
    \small
    \centering
    \footnotesize
    \includegraphics[clip, trim=0cm 9cm 2cm 0.4cm, width=\linewidth]{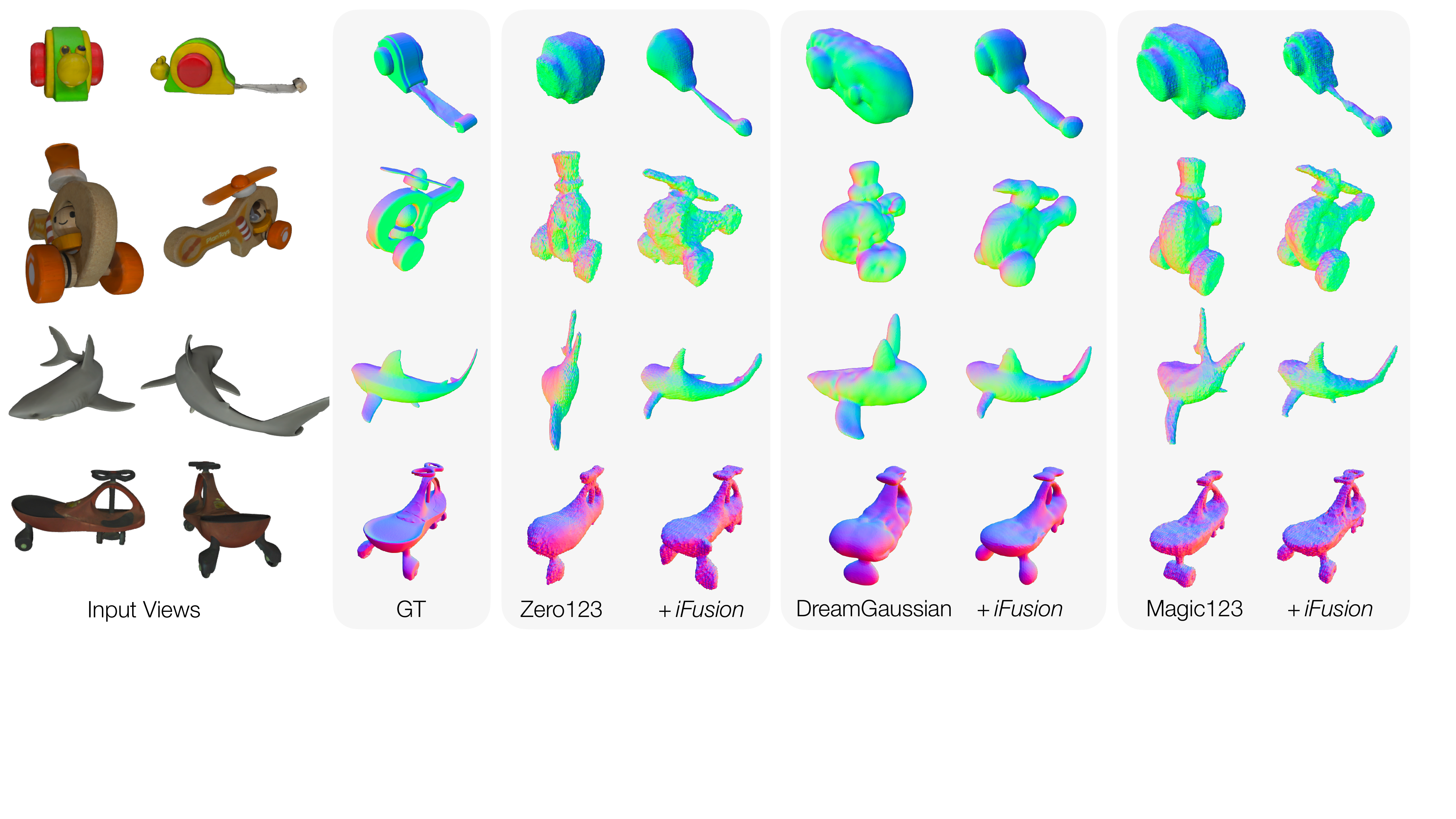}
    \vspace{-5mm}
    \caption{\textbf{Qualitative comparison of surface reconstruction.}
    It is clear that \ours significantly enhances existing reconstruction methods including Zero123-SDS~\cite{zero123},~DreamGaussian~\cite{dreamgaussian}, and~Magic123~\cite{magic123}, by adding an additional view without the camera pose.
    }
    \label{fig_exp_recon}
\end{figure*}

%% file: table/exp_recon.tex
\begin{table*}
\fontsize{8.8}{12}\selectfont
  \caption{\textbf{Evaluation results on 3D reconstruction.} Strong single-view reconstruction baselines are improved by \ours consistently.
  }
  \label{exp_recon}
  \centering
  \begin{tabular}{l cc cc}
    \toprule

    \multirow{2}{*}{Method} & \multicolumn{2}{c}{GSO~\cite{gso}} & \multicolumn{2}{c}{OO3D~\cite{omniobject3d}} \\
    \cmidrule(lr){2-3} \cmidrule(lr){4-5}
      & Chamfer Dist. ($\times 10^3)\downarrow$ & Volume IoU (\%) $\uparrow$ & Chamfer Dist. ($\times 10^3)\downarrow$ & Volume IoU (\%) $\uparrow$ \\
    
    \midrule
    
    Point-E~\cite{point_e}  & 6.414 & 18.92 & 6.766 & 19.83 \\
    Shape-E~\cite{shap_e}  & 5.839 & 29.00 & 6.086 & 29.02\\
    \midrule
    One-2-3-45~\cite{one2345}  & 7.173 & 28.77 & 5.424 & 43.75 \\
    $\;\;$+ \ours &  \underline{6.359} & \underline{31.68} & \underline{4.739} & \underline{48.32} \\
    Zero123-SDS~\cite{zero123}  & 6.456 & 33.63 & 5.676 & 45.90 \\
    $\;\;$+ \ours &  \underline{4.178} & \underline{39.73} & \underline{3.293} & \underline{56.36} \\
    DreamGaussian~\cite{dreamgaussian}  & 4.728 & 35.35 & 4.298 & 44.35 \\
    $\;\;$+ \ours & \underline{3.977} & \underline{42.07} & \underline{2.947} & \underline{57.58} \\
    Magic123~\cite{magic123}  & 4.839 & 39.46 & 3.842 & 53.69 \\
    $\;\;$+ \ours &  \textbf{3.076} & \textbf{46.70} & \textbf{2.682} & \textbf{60.31} \\
    
    \bottomrule

  \end{tabular}
  \vspace{-8pt}

\end{table*}

%% file: table/exp_ab_pose.tex
\begin{table}
\fontsize{9}{12}\selectfont
  \caption{Ablation of the \textbf{number of initial poses} for pose estimation on GSO~\cite{gso}.}
  \vspace{-4pt}
  \label{exp_ab_pose}
  \centering
  \newcolumntype{Y}{>{\centering\arraybackslash}X}
  \begin{tabularx}{\columnwidth}{l c *{3}{Y} c}
    \toprule
    
    & \multirow{2}{*}{$n$ poses}  & \multicolumn{3}{c}{Recall $\uparrow$} & \multirow{2}{*}{Time (s) $\downarrow$} \\
    \cmidrule{3-5} & & 5$^\circ$ & 10$^\circ$ & 20$^\circ$ \\

    \midrule

    (a) & $1$ & 33.07 & 36.21 & 38.36 & 22.30 \\
    (b) & $2$ & 60.57 & 69.14 & 73.07 & 38.51 \\
    (c) & $4$ & 74.79 & 84.29 & 88.57 & 70.59 \\
    (d) & $8$ & \textbf{78.21} & \textbf{88.93} & \textbf{92.43} & 133.73 \\
    
    \bottomrule
  \end{tabularx}

\end{table}

\begin{table}
\fontsize{9}{12}\selectfont
  \caption{Ablation of \textbf{$\mathbf{t}$ annealing} for pose estimation on GSO~\cite{gso}.}
  \vspace{-4pt}
  \label{exp_ab_pose_t}
  \centering
  \newcolumntype{Y}{>{\centering\arraybackslash}X}
  \begin{tabularx}{\columnwidth}{l c c *{3}{Y}}
    \toprule

    & \multirow{2}{*}{$n$ poses} & \multirow{2}{*}{$t$ annealing} & \multicolumn{3}{c}{Recall $\uparrow$} \\
    
    \cmidrule{4-6} & & & 5$^\circ$ & 10$^\circ$ & 20$^\circ$ \\

    \midrule
    
    (a) & $4$ & - & 48.61 & 56.67 & 61.39 \\
    (b) & $4$ & \checkmark & \textbf{74.79} & \textbf{84.29} & \textbf{88.57} \\
    
    \bottomrule
  \end{tabularx}
  \vspace{-6pt}
  
\end{table}

%% file: sec/5_related.tex
\section{Related Work}
\label{sec:related}

\paragraph{Few-shot NeRFs} \label{sec_related_nerf}
Neural Radiance Fields~(NeRFs)~\cite{nerf} have revolutionized 3D modeling with its powerful representations and high-fidelity render quality, but struggling under insufficient views. Follow-up works introduced regularizations to stabilize training~\cite{infonerf, freenerf, regnerf}, or prior models for auxiliary 3D reasoning~\cite{pixelnerf, dietnerf, mvsnerf, ibrnet}. Nevertheless, the dependency on precise camera poses remains an issue, as~\citet{barf} showed that inaccurate poses, which often arise in pose estimation using a limited number of views, lead to degraded performance.

\vspace{-10pt}
\paragraph{Diffusion for 3D Generation} 
Diffusion models~\cite{dpm, ddpm, ncsn} have emerged as the leading visual generative models. They generate visually plausible images from various input conditions~\citep{fan2023frido,yang2023law,xue2023freestyle,meng2022sdedit,gafni2022make,li2023gligen} and customize or edit existing photos with diverse controlling signals~\citep{zhao2023uni,zhang2023adding,avrahami2023spatext,gal2023an,ruiz2023dreambooth,qin2023unicontrol}. Promising results have been achieved in 3D generation as well, spanning various representations such as point-clouds~\cite{diff_pcl, lion, pvd},~voxel grids~\cite{pvd, diffrf}, and~tri-planes~\cite{triplane, renderdiffusion, nerfdiff}; however, they are constrained by the limited diversity of 3D datasets, \eg, ShapeNet~\cite{shapenet}. To overcome the data scarcity, researchers utilize pre-trained 2D diffusion models~\cite{imagen, ldm} for text-to-3D generation~\cite{dreamfusion, magic3d, fantasia3d, prolificdreamer}, and further extend them for single-view reconstruction~\cite{realfusion, make_it_3d, one2345, magic123, zero123, dreamgaussian}, where the diffusion model \textit{``dreams up"} unobserved views.
However, single-view methods diverge from real-world reconstruction scenarios --- the target object needs to be accurately reconstructed, not over-imagined. Although several methods propose to include additional views, accurate camera poses are still assumed~\cite{sparsefusion, viewset_diffusion, genvs, holodiffusion}.

\vspace{-10pt}
\paragraph{Pose-free Reconstruction} 

To recover the unknown camera poses from sparse views, recent studies have explored learnable pose estimation, either by directly regressing the pose~\cite{relpose, relposepp, forge} or through iterative refinement~\cite{sparsepose, posediffusion}. The estimated poses can then be utilized for reconstruction~\cite{barf, ners, forge}. Notably, FORGE~\cite{forge} combines the two stages to achieve pose-free reconstruction but lacks robustness for intricate geometry and is sensitive to lighting. A recent follow-up, LEAP~\cite{leap}, eliminates pose estimation by employing DINOv2~\citep{oquab2023dinov2} for feature mapping, showing improved generalization but struggling at unseen regions. In contrast, our solution excels in these scenarios, empirically proving its value in extreme few-shot situations.

%% file: table/exp_ab_nvs.tex
\begin{table}
\fontsize{8.9}{12}\selectfont
  \caption{\textbf{Ablation of novel view synthesis} on GSO~\cite{gso}. Multi-view conditioning and LoRA~\cite{lora} finetuning are validated. Increased views also improve the scores.}
  \vspace{-4pt}
  \label{exp_ab_nvs}
  \centering
  \begin{tabularx}{\columnwidth}{l c l c c c}
    \toprule
    
     & $n$ views & Strategy & LoRA & PSNR $\uparrow$ & LPIPS $\downarrow$ \\

    \midrule

    (a) & 1 & - & - & 15.40 & 0.184 \\
    (b) & 2 & closest-view & - & 16.19 & 0.169 \\
    (c) & 2 & multi-view & - & 17.30 & 0.149 \\
    (d) & 2 & multi-view & \checkmark & 18.73 & 0.121 \\
    (e) & 4 & multi-view & \checkmark & \textbf{21.32} & \textbf{0.092}\\
    
    \bottomrule
  \end{tabularx}
\end{table}

%% file: table/exp_ab_recon.tex
\begin{table}
\fontsize{8.7}{12}\selectfont
  \caption{\textbf{Ablation of 3D reconstruction} on OO3D~\cite{omniobject3d} based on DreamGaussian~\cite{dreamgaussian}. MVC + LoRA achieves the best result.}
  \vspace{-4pt}
  \label{exp_ab_recon}
  \centering
  \begin{tabularx}{\columnwidth}{l c c c c c}
    \toprule
    
     & $n$ views & MVC & LoRA & Chamfer Dist. $\downarrow$ & IoU $\uparrow$ \\

    \midrule

    (a) & 1 & - & - & 4.298 & 44.35 \\
    (b) & 2 & - & - & 3.427 & 53.04 \\
    (c) & 2 & \checkmark& - & 3.241 & 54.16 \\
    (d) & 2 & \checkmark & \checkmark & \textbf{2.947} & \textbf{57.58} \\
    
    \bottomrule
    
  \multicolumn{5}{l}{\footnotesize *Chamfer distance measured by $\times 10^3$ and IoU in (\%)} \\
  \end{tabularx}
  \vspace{-6pt}
\end{table}

%% file: sec/6_conclusion.tex
\section{Conclusion}
\label{sec:conclusion}

We propose \ours, a framework that reconstructs 3D objects without requiring poses, leveraging a large-scale pre-trained diffusion model as a prior. Given a few unposed images, we begin with inverting the diffusion for gradient-based pose optimization. The estimated poses, in turn, enhance the novel view synthesis diffusion model through multi-view fine-tuning and conditioning. Finally, by combining the estimated poses and the refined diffusion model, we demonstrate how \ours achieves pose-free reconstruction. Experimental results show that our solution outperforms strong baselines on three key tasks: pose estimation, novel view synthesis, and 3D reconstruction.

%% file: sec/7_ack.tex
\section*{Acknowledgement}

This work is supported in part by the National Science and Technology Council~(NSTC 111-2634-F-002-022).
The views and opinions expressed in this paper are solely those of the authors and do not necessarily represent the official policies or positions of their affiliations or funding agencies.

%% file: sec/a_impl.tex
\section{Implementation Details}

\subsection{Hyper-parameters}
We employ Adam~\cite{adam} as the optimizer for both pose estimation and sparse-view fine-tuning. In the case of pose estimation, we optimize the initial poses for~100 steps with an initial learning rate of~0.1. The learning rate is dynamically reduced if the L2 loss stops decreasing, handled by the \textit{ReduceLROnPlateau} scheduler from PyTorch~\cite{pytorch}. Specifically, we set the reduction factor to~0.6 and the patience to~10. Afterward, in the sparse-view fine-tuning stage, the model is fine-tuned for~30 steps, with the learning rate annealed from~$10^{-3}$ to~$10^{-4}$ and the rank of the injected LoRA parameter set to~12. This process takes approximately 30 seconds on a single Nvidia 3090 GPU with a batch size of~16. For 3D reconstruction, we follow the default hyper-parameters of each reconstruction method, \ie, One2345~\cite{one2345},~Zero123-SDS~\cite{zero123},~Magic123~\cite{magic123}, and~DreamGaussian~\cite{dreamgaussian}, when combining with \ours. Please refer to their official implementations for details.

\subsection{Dataset Collection}
We use Pyrender to render images for evaluation.\footnote{\url{https://github.com/mmatl/pyrender}}
Following~\citet{zero123}, the transformation is defined using the spherical coordinate system with~$\theta$,~$\phi$, and~$r$ representing the elevation angle, azimuth angle, and distance towards the center, respectively. In practice, we sample camera viewpoints on the unit sphere with~$\theta \in [\pi/4, 3\pi/4]$, $\phi \in [0, 2\pi]$ and $r$ is uniformly sampled in the interval of~$[1.2, 2.0]$. The field of view of the perspective camera is set to~49.1$^\circ$. All images are rendered in the resolution of~512$\times$512 with transparent background.

%% file: sec/b_qual.tex
\section{Qualitative Results}

\input{fig/supp_ambi}

To further corroborate the effectiveness of our proposed pose estimation strategy described in~\cref{sec:method:pose}, we present additional qualitative visualization in~\cref{fig_supp_pose}. These results complement the findings presented in~\cref{fig_exp_pose} of our main manuscript. They also support our assumption that the learned understanding of diverse objects in Zero123~\cite{zero123} can be leveraged for other tasks, such as pose estimation. Moreover, examples illustrating the single-view ambiguity, taken from the Blender dataset~\cite{nerf}, are shown in~\cref{fig_supp_ambi}. These instances motivated us to fine-tune and condition the model with registered multi-views.

In~\cref{fig_supp_recon}, we showcase additional comparisons on 3D reconstruction. These results complement~\cref{fig_exp_recon} of our main manuscript, underscoring the efficacy of our proposed framework in achieving faithful reconstruction by considering one extra view without requiring known camera poses.

%% file: fig/supp_ambi.tex
\begin{figure}[t]
    \small
    \centering
    \footnotesize
    \includegraphics[clip, trim=0cm 12.5cm 40cm 0cm, width=\columnwidth]{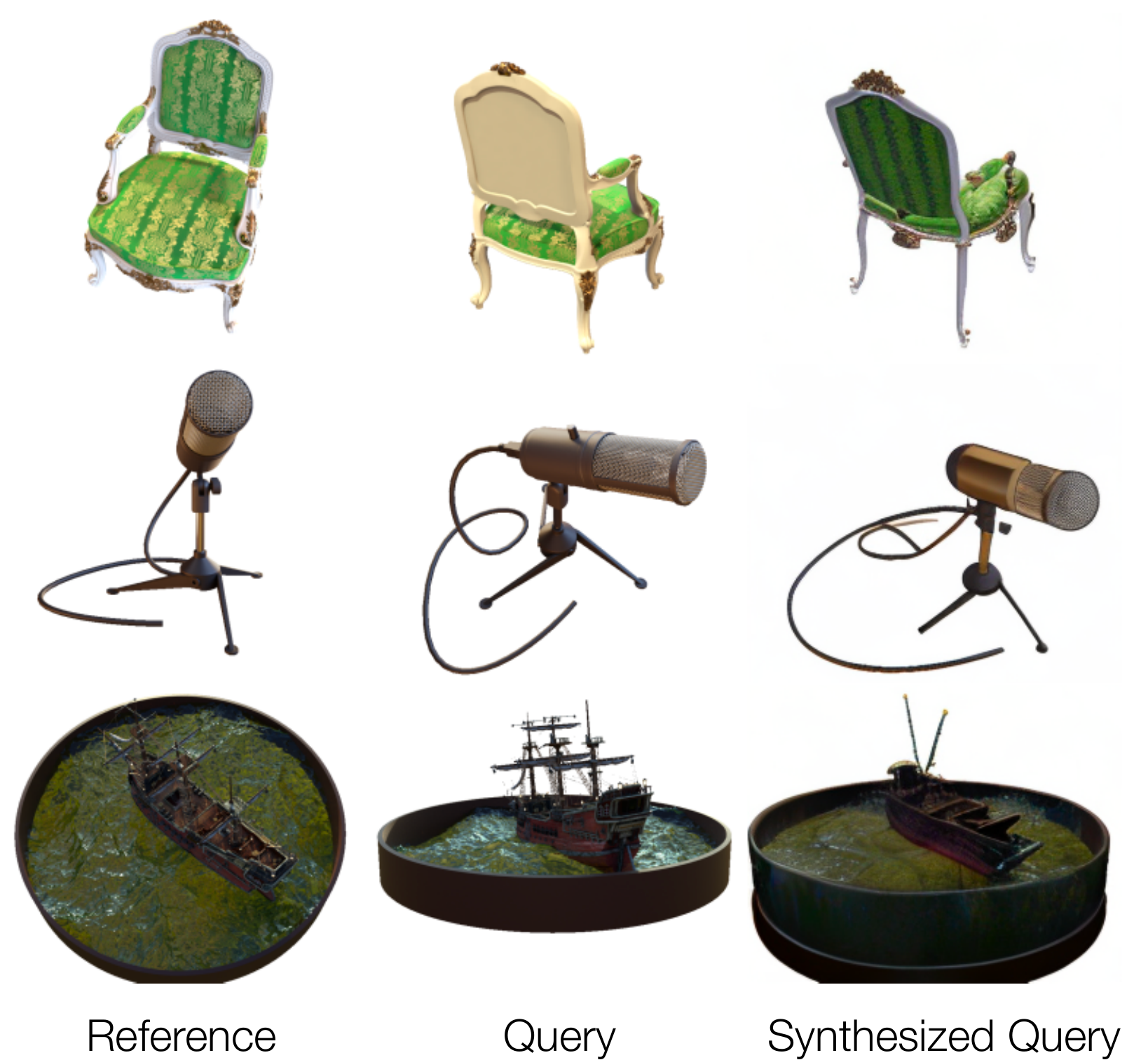}
    \caption{\textbf{Single-view ambiguity.} We show the reference view, query view, and the synthesized query view given $c(x^r, T^r_q)$. It is observed that, while the model can generate reasonable novel views, there is a gap between the model's understanding and the actual object, arising from single-view ambiguity. This prompts us to condition the model with additional views to mitigate this issue.
    }
    \label{fig_supp_ambi}
\end{figure}

%% file: sec/c_limitation.tex
\section{Limitations and Future Works} 

While our methods deliver highly accurate camera poses, our pose estimation run time is higher than feed-forward-based methods, \eg, RelPose++~\cite{relposepp}. This is attributed to the optimization nature of our approach, which involves back-propagation for updating the poses. Moreover, when we fine-tune Zero123~\cite{zero123} on estimated poses and additional input views, it is worth noting that Zero123, originally adapted from the 2D-based Stable Diffusion (SD), lacks complete 3D awareness. This structural limitation prevents it from generating multi-views with consistency. However, our framework holds potential for integration with other diffusion-based novel view synthesizers~\cite{syncdreamer, wonder3d} that enforce consistency by incorporating 3D-aware modules onto SD.

%% file: fig/supp_pose.tex
\begin{figure*}[hbt!]
    \small
    \centering
    \footnotesize
    \includegraphics[clip, trim=0cm 0cm 35.5cm 0cm, width=0.9\linewidth]{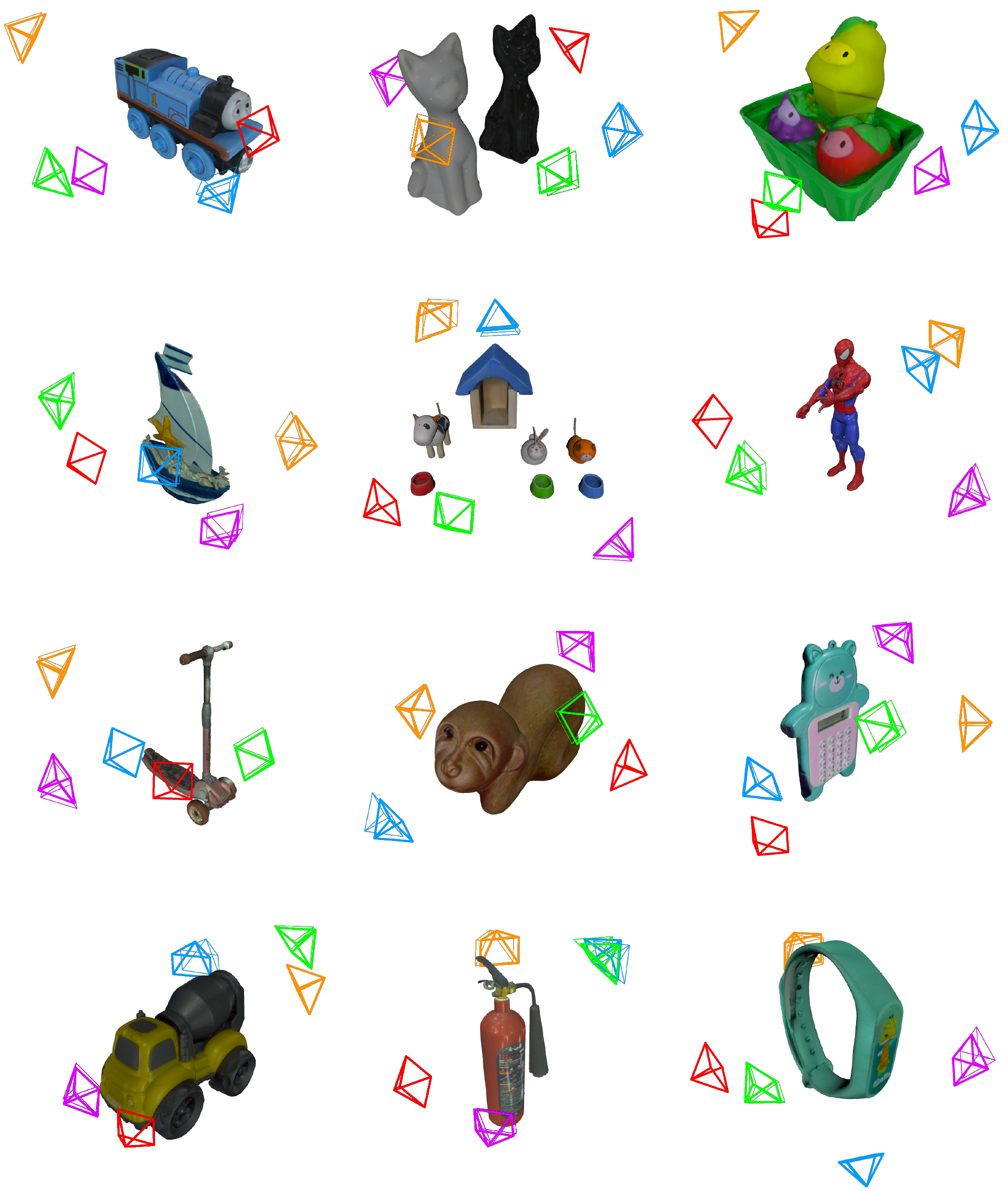}
    \caption{\textbf{More qualitative results on pose estimation.} The predicted poses~(thin) and their corresponding ground truth~(bold), are plotted in the same color, while the \textcolor{red}{reference views} are plotted in red. We confirm that \ours effectively exploits the robust understanding of diverse objects in Zero123~\cite{zero123} acquired from Objaverse~\cite{objaverse}.
    }
    \label{fig_supp_pose}
\end{figure*}

%% file: fig/supp_recon.tex
\begin{figure*}[hbt!]
    \small
    \centering
    \footnotesize
    \includegraphics[clip, trim=0cm 1.5cm 0cm 0cm, width=\linewidth]{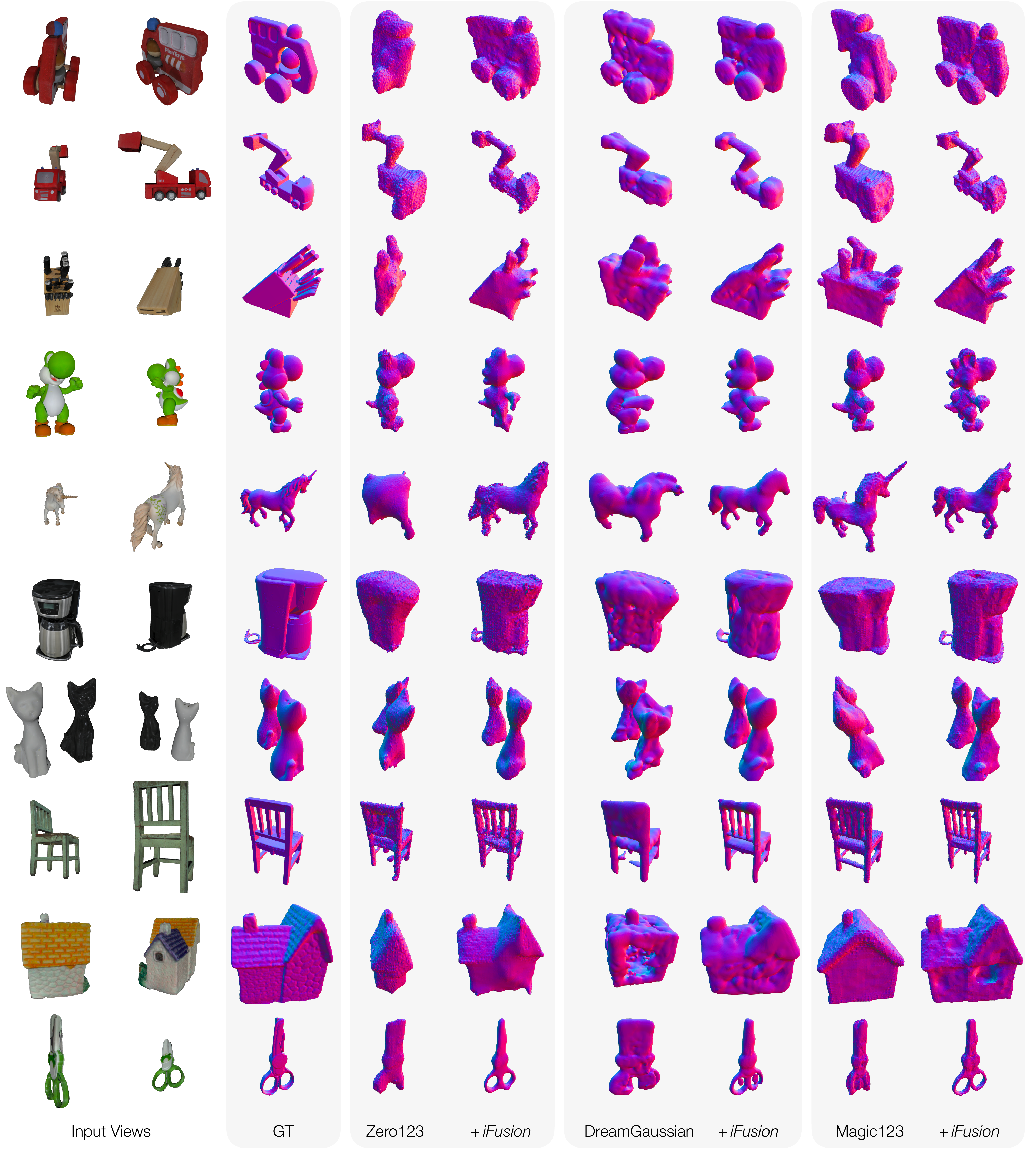}
    \caption{\textbf{More qualitative comparisons on surface reconstruction.} We integrate \ours with Zero123-SDS~\cite{zero123},~DreamGaussian~\cite{dreamgaussian}, and~Magic123~\cite{magic123} to perform pose-free reconstruction given sparse views. The results indicate that our method operates as an effective add-on, consistently enhancing existing single-view reconstruction methods.
    }
    \label{fig_supp_recon}
\end{figure*}